\documentclass[a4paper, 10pt, conference]{ieeeconf}      

\IEEEoverridecommandlockouts                              
\makeatletter
\let\NAT@parse\undefined
\makeatother

\usepackage[dvipsnames]{xcolor}

\newcommand*\linkcolours{ForestGreen}

\usepackage{times}
\usepackage{graphicx}
\usepackage{amssymb}
\usepackage{gensymb}
\usepackage{amsmath}
\usepackage{breakurl}

\usepackage{url,hyperref}
\hypersetup{
colorlinks,
linkcolor=\linkcolours,
citecolor=\linkcolours,
filecolor=\linkcolours,
urlcolor=\linkcolours}

\usepackage{algorithm}
\usepackage{algorithmic}

\usepackage[labelfont={bf},font=small]{caption}
\usepackage[none]{hyphenat}

\usepackage{graphicx}

\usepackage{mathtools, cuted}

\usepackage[noadjust, nobreak]{cite}

\usepackage{tabularx}
\usepackage{amsmath}
\usepackage{bm}

\usepackage{float}

\usepackage{pifont}

\usepackage{xspace}
\newcommand{\eva}{\textit{eva}\xspace}
\newcommand{\odcorr}{\text{ODCorr}\xspace}
\newcommand{\repourl}{\url{https://github.com/kaiko-ai/towards_large_pathology_fms}\xspace}

\newcolumntype{Y}{>{\centering\arraybackslash}X}

\usepackage[]{placeins}

\usepackage{placeins}

\usepackage{tikz}

\usepackage[framemethod=tikz]{mdframed}

\usepackage{afterpage}

\usepackage{stfloats}

\usepackage{atbegshi}
\newcommand{\handlethispage}{}
\newcommand{\discardpagesfromhere}{\let\handlethispage\AtBeginShipoutDiscard}
\newcommand{\keeppagesfromhere}{\let\handlethispage\relax}
\AtBeginShipout{\handlethispage}

\usepackage{comment}

\usepackage{tabularx}               
\usepackage{multirow}               
\usepackage{booktabs}               
\usepackage{colortbl}               
\usepackage{ulem}                   
\usepackage{hyperref}               
\usepackage{url}                    
\usepackage{arydshln}               
\usepackage{hhline}                 

\title{\LARGE \bf
Towards Large-Scale Training of Pathology Foundation Models
}

\author{%
kaiko.ai,
Nanne Aben,
Edwin D. de Jong, 
Ioannis Gatopoulos,
Nicolas K\"anzig,
Mikhail Karasikov,\\
Axel Lagré,
Roman Moser,
Joost van Doorn,
Fei Tang\textsuperscript{\textdagger}
\thanks{\textsuperscript{\textdagger}Correspondence E-mail: fei@kaiko.ai. Other authors are ordered alphabetically.}
\\{\texttt{kaiko.ai}}
}

\begin{document}

\maketitle
\thispagestyle{empty}
\pagestyle{empty}

%
\noindent
\begin{abstract}
\noindent
Driven by the recent advances in deep learning methods and, in particular, by the development of modern self-supervised learning algorithms, increased interest and efforts have been devoted to build foundation models (FMs) for medical images.
In this work, we present our scalable training pipeline for large pathology imaging data, and a comprehensive analysis of various hyperparameter choices and training techniques for building pathology FMs.
We release and make publicly available the first batch of our pathology FMs\footnote{\repourl} trained on open-access TCGA whole slide images, a commonly used collection of pathology images.
The experimental evaluation shows that our models reach state-of-the-art performance on various patch-level downstream tasks, ranging from breast cancer subtyping to colorectal nuclear segmentation.
Finally, to unify the evaluation approaches used in the field and to simplify future comparisons of different FMs, we present an open-source framework\footnote{\url{https://github.com/kaiko-ai/eva}} designed for the consistent evaluation of pathology FMs across various downstream tasks.

\end{abstract}

\section{INTRODUCTION}
\noindent

Pathology images contain a wealth of information about patient health status, and deep learning-based methods have become increasingly capable of automatically extracting vital information about a patient's condition from these images \cite{campanella2019clinical,lu2021data,echle2021deep,tellez2018whole,bulten2020automated,litjens2017survey,van2021deep,dimitriou2019deep}. While expert human pathologists are able to detect certain subtle patterns from Whole Slide Images (WSIs), such as Micro-Satellite Instability \cite{Hildebrand2021,Zhu2022,saillard2021self}, AI methods are increasingly able to detect ever subtler patterns that even expert pathologists are unable to detect visually. For example, Machine Learning (ML) models have been trained to predict molecular biomarkers  \cite{mccaw2024machine,el2024regression} and RNA expression levels from pathology images \cite{schmauch2020deep}.

One of the clearest findings from the past decade of machine learning research is that increasing training dataset size and variety is a primary driver of increased model performance. This trend can be traced from AlexNet \cite{krizhevsky2012imagenet} and the subsequent development of convolutional neural network architectures enabled by ImageNet, e.g.~\cite{simonyan2014very,he2016deep,he2017mask}. Self-Supervised Learning (SSL) then facilitated exploiting seemingly unlimited further increases in dataset magnitude, up to the internet-scale datasets that are used to train current Large Language Models (LLMs) such as GPT-4 \cite{achiam2023gpt}.

Hospitals worldwide are collecting data at an unprecedented scale. With this data, new challenges and opportunities arise: How can AI methods best be employed to extract and make use of the wealth of medically relevant information contained in large-scale medical imaging datasets?

The work of Campanella et al.~\cite{campanella2019clinical} was pivotal in pathology machine learning for demonstrating that  WSI-level supervision (weak labels) is sufficient to effectively train high-quality pathology models given sufficiently large datasets (approx.~44k WSIs), thus obviating the need for painstaking and time-consuming annotation efforts of pathologists.

That early work and the simultaneous advancement of Self-Supervised Learning (SSL)~\cite{DBLP:journals/corr/abs-2002-05709-simclr,DBLP:journals/corr/abs-2103-00020-clip,barlow-twins-DBLP:journals/corr/abs-2103-03230,swav-DBLP:journals/corr/abs-2006-09882,caron2021emerging,oquab2023dinov2} paved the way for the recent series of pathology SSL models trained on increasingly large datasets. To the best of our knowledge, the earliest work that applied SSL to pathology images is ``Self-supervision closes the gap between weak and strong supervision in histology" \cite{dehaene2020self}. HIPT \cite{chen2022scaling} then used a hierarchical setup where Vision Transformers (ViTs) are trained with the self-supervised DINO \cite{caron2021emerging} algorithm on 10,678 FFPE (formalin-fixed, paraffin-embedded) WSIs from TCGA~\cite{Chang:2013aa}. Lunit~\cite{Kang_2023_CVPR} trained on 19M patches from the full set of 21k TCGA WSIs. Phikon (Owkin)~\cite{Filiot2023.07.21.23292757} is an iBOT ViT-Base model (80M parameters) trained on over 40M patches from 6k WSIs.

Chen et al.~\cite{chen2023generalpurpose} trained the UNI model, and it is the first work to use an order of magnitude more WSIs than TCGA. The UNI model was trained on over 100M patches from over 100k diagnostic H\&E WSIs across 20 tissue types and evaluated on 33 representative clinical pathology tasks of varying difficulty. Campanella et al.~\cite{campanella2023computational} claimed to have collected the largest pathology dataset at the time (Oct 2023) and trained on 3B patches from over 423k WSIs, comparing pre-training of vision transformer models using a masked autoencoder (MAE) vs.~DINO. However, these numbers are surpassed remarkably by the 632M parameter Virchow model trained on 1.5M WSIs~\cite{vorontsov2024virchow}. 

In collaboration with the Netherlands Cancer Institute (NKI-AvL), we are building one of the first large-scale pathology FM on a European cohort. Together with NKI, we intend to significantly increase the dataset size to 1M and beyond, also in partnership with other cancer centers. Working with data at this scale brings both infrastructure and architectural challenges, which we address in this work.


Here, we present our scalable pipeline for training and evaluating foundation models (FMs) on large pathology imaging data. For seamless training, we developed Online Patching, a technique for high-throughput loading of arbitrary image patches cropped from large WSIs residing in blob storage (described in Section~\ref{sec:online_patching}). With this, we trained vision transformers of various sizes using the DINO and DINOv2 SSL algorithms on TCGA WSIs, an open-access pathology image dataset, which has been widely used in the community for training pathology FMs. We present our best-to-date FMs trained on TCGA and compare them to state of the art. The evaluation shows that our FMs perform on par or better than the existing state-of-the-art FMs on most downstream tasks (Section \ref{results-overview}).

We also present an experimental study of various hyperparameter and design choices for training pathology FMs, such as the model initialization strategy (Section \ref{experiment-scratch-vs-pretrained}), effects of mixing different magnifications (Section \ref{experiment-mpps}), and effects of data sizes (Section \ref{experiment-data-sizes}), which may be of interest to other practitioners in the field of medical machine learning and computational pathology to guide the future development of pathology FMs.

To aid the evaluation of FMs, we introduce a new unsupervised metric that can be used to compare models of different sizes and show that it correlates well with downstream performance and, hence, can be a valuable addition to supervised metrics (Section \ref{sec:unsupervised metric} and \ref{supp:odcorr}). 

Lastly, we developed \eva (described in Section \ref{section:eva}), an open-source framework for evaluating FMs on clinically relevant downstream tasks in a straightforward and unified way. 
We will be extending \eva with more downstream tasks in the future and invite other practitioners in the field to contribute more downstream tasks to \eva in order to build a standardized evaluation workflow and to ensure the evaluation results are comparable across different studies.

\section{Results and discussion}
\begin{table*}[t]
\caption[Patch-level benchmarks]{Linear probing evaluation of FMs on patch-level downstream datasets. We compare the performance of a randomly initialized model (ViT-S16 {\color[HTML]{9B9B9B} \textit{(rand.)}}), FMs trained on ImageNet (above the dashed line), and the pathology FMs (below the dashed line).
The evaluation was performed with \eva.
For BACH, CRC, MHIST, PCam, and TP53, the numbers represent the balanced accuracy averaged over 5 linear probing runs (the respective standard deviations can be found in the extended version of Table~\ref{tab:results-appendix}). For CoNSeP, we report the DICE score on the foreground pixels. The best results are highlighted in bold. (*) For Virchow, the model weights were not publicly available. The values were taken from \cite{vorontsov2024virchow}, where they were computed in a different setup and may not be directly comparable to the other values.
(**) For TP53 and CoNSeP, the evaluation was done separately but will soon be supported in \eva as well.
}
\centering
\scalebox{1.1}{
\begin{tabular}{lccccccccc}
\textbf{Model} & \textbf{Training data} & \textbf{BACH} & \textbf{CRC} & \textbf{MHIST}  & \textbf{PCam} & \textbf{TP53}** & \textbf{CoNSeP}**\\[0.1ex]
\hline & \\[-1.7ex]
ViT-S16 \color[HTML]{9B9B9B} \textit{(rand.)} & None & 0.410 & 0.617 & 0.501  & 0.728 & 0.500 & 0.583\\ 
DINO ViT-S16 \cite{caron2021emerging} & ImageNet & 0.695 & 0.935 & {0.831} & 0.849 & 0.519 &  0.611\\ 
DINO ViT-B8 \cite{caron2021emerging} & ImageNet & 0.710 & 0.939 & 0.814 & 0.856 & 0.548 & 0.710\\[1ex]
\arrayrulecolor{gray} \hdashline[1.5pt/1.5pt] \\[-1.5ex]
Lunit \cite{Kang_2023_CVPR} & TCGA (21k WSIs) & 0.801 & 0.934 & 0.768 & 0.895 & 0.571 & 0.654\\ 
Phikon \cite{Filiot2023.07.21.23292757} & TCGA (6k WSIs) & 0.725 & 0.935 & 0.777 & {0.915} & 0.630 & 0.666\\
DINO ViT-S16 \color[HTML]{9B9B9B} \textit{(ours)} & TCGA (29k WSIs) & 0.797 & 0.943 & {0.828} & 0.893 & 0.633 & 0.649\\ 
DINO ViT-S8 \color[HTML]{9B9B9B} \textit{(ours)} & TCGA (29k WSIs) & 0.834 & 0.946 & \textbf{0.832} & 0.887 & 0.621 & {0.724}\\ 
DINO ViT-B16 \color[HTML]{9B9B9B} \textit{(ours)} & TCGA (29k WSIs) & 0.810 & \textbf{0.960} & 0.826 & 0.898 & 0.651 & 0.658\\ 
DINO ViT-B8 \color[HTML]{9B9B9B} \textit{(ours)} & TCGA (29k WSIs) & {0.865} & {0.956} & 0.809 & \textbf{0.921} & \textbf{0.659} & \textbf{0.741}\\ 
DINOv2 ViT-L14 \color[HTML]{9B9B9B} \textit{(ours)} & TCGA (29k WSIs) & \textbf{0.870} & 0.930 & 0.809 & 0.898 & {0.656} & 0.679\\ 
Virchow \cite{vorontsov2024virchow} & Private (1.5M WSIs) & n/a & \color[HTML]{9B9B9B} 0.962\mbox{*} & \color[HTML]{9B9B9B} 0.830\mbox{*} & \color[HTML]{9B9B9B} 0.933\mbox{*}& n/a & n/a \\
\bottomrule
\end{tabular}}
\label{tab:results}
\end{table*}

\subsection{Training state-of-the-art FMs with online patching}
\label{results-overview}
In most existing work, a patch dataset is typically pre-constructed from WSIs offline before training, which results in a fixed set of patches.
Moreover, for larger datasets, this approach becomes inefficient for the following reasons: 1)~For every new patch extraction strategy, the dataset has to be re-created from scratch, which is costly and time-consuming. 2) In addition, every time a patch dataset is created, it requires a large storage space overhead, which makes dynamic patch sampling and experiments with sampling strategies practically impossible.

To address these issues, we developed {\it Online Patching} (see Methods, Section~\ref{sec:online_patching}), a method that allows for the online high-throughput extraction of patches of arbitrary size and resolution from WSIs residing in blob storage.
Not only does online patching improve data processing efficiency, but it also introduces a key difference to the offline patching approach: the patches are created dynamically.
As a result, dynamic patch sampling strategies can be seamlessly incorporated into the training procedure.

Even for a single WSI with $10^5\times 10^5$ pixels, there can be up to $10^{10}$ distinct sampled patches.
With Online Patching, almost every sampled patch is new because it is sampled at a random position. Thus, the number of distinct patches our models have seen during training is typically much larger than for other models trained with offline patching. Moreover, this number grows with the number of training epochs. On the other hand, many of the patches, despite being unique, do overlap with many other neighboring patches.
It is currently unclear how the number of unique patches and their similarity impact the performance of the trained FM. In our experiments, we implicitly show that sampling all patches at random coordinates does not have a negative impact on the performance of the trained FM, and we leave a more comprehensive analysis for future work.

Using online patching, we trained several vision transformer models of different sizes using both the DINO and DINOv2 algorithms on the whole set of 29k open-access Flash-Frozen (FF) and FFPE diagnostic tissue slides from TCGA. We evaluated the resulting models and compared them to the existing state-of-the-art models.

For training our FMs, we mainly followed the original recipes from DINO~\cite{caron2021emerging} and DINOv2~\cite{oquab2023dinov2}. More specifically, we deviate from the original recipes in the following: 1) We start from the models pre-trained on ImageNet published in~\cite{caron2021emerging} and~\cite{oquab2023dinov2}, respectively; 2) Our FMs are trained on patches extracted from TCGA WSIs at different magnification levels; 3) We use fewer GPUs and, consequently, a smaller global batch size. (For example, for ViT-B8, the original config from the DINO repository specified 176 GPUs, while we only use 8 GPUs.) We used the linear and square root scaling law for the learning rate in DINO and DINOv2, respectively. 
For more details, see Section~\ref{subsection:pretraining setup}.

The overview and the performance of our best models is presented in Table~\ref{tab:results}, descriptions for each downstream task can be found in Data section in \ref{section:data}. The ViT-S16 model we trained is comparable to the state-of-the-art models of similar size on all considered downstream tasks. Notably, on BACH, CRC, and MHIST, it achieves higher accuracy than the larger Phikon model, which is a ViT-B16 model trained with iBOT.

Similar to what is reported in DINO~\cite{caron2021emerging}, reducing the patch size used in the vision transformer considerably improves the model performance. This is especially prominent in the segmentation task, where the top two performing models are ViT-B8 and ViT-S8 surpassing the next in line, ViT-L14, by a large margin, despite the fact that the ViT-L14 has more parameters and was trained with DINOv2 with the additional patch-level objectives.

On the other hand, unlike what was observed in DINO~\cite{caron2021emerging}, scaling up the model size has shown on TCGA a limited impact on the performance (e.g., the performance on PCam changed from 0.893 for ViT-S16 to 0.921 for ViT-B8, and to 0.898 for ViT-L14.). We hypothesize that the impact could be limited for two possible reasons. First, the performance on some downstream tasks might have reached its maximum, and hence, we can no longer observe any differences between the FMs. Secondly, the larger models might have not reached their full capacities, either because the effective data size of the TCGA images is too small due to the limited diversity or because the larger models were not trained long enough. Virchow~\cite{vorontsov2024virchow}, for example, achieved superior performance on PCam and CRC with a ViT-H/14 trained on almost two orders of magnitude more WSIs.

\begin{table*}[t]
\caption[Magnification ablation study]{Magnification ablation study: evaluating downstream benchmark performance through different patch magnifications in pre-training phase. All results were generated using  \textit{eva}. All runs have, on average, a standard deviation of (±0.002). (*) The images from BACH were downsampled from an mpp of 0.42\,µm/px (20$\times$) to 2.88\,µm/px (3.47$\times$).
}
\centering
\scalebox{1.1}{
\begin{tabular}{lcccccccc}
\textbf{Downstream task} & \textbf{40\texttimes} & \textbf{20\texttimes} & \textbf{10\texttimes} & \textbf{5\texttimes} & \textbf{\{20,\,40\}\texttimes}  & \textbf{\{5,\,10,\,20\}\texttimes}  & \textbf{\{5,\,10,\,20,\,40\}\texttimes}\\[0.1ex]
\hline & \\[-1.7ex]
BACH (3.47\texttimes
)* & 0.639 & 0.685 & 0.659 & 0.679 & 0.689 & 0.683 & \textbf{0.753} \\ 
CRC (20\texttimes) & 0.935 & 0.945 & 0.939 & 0.927 & 0.942 & 0.944 & \textbf{0.947}\\ 
MHIST (5\texttimes) & 0.744 & 0.746 & 0.648 & 0.710 & 0.746 & 0.744 & \textbf{0.771} \\ 
PCam/val (10\texttimes) & 0.879 & \textbf{0.898} & 0.873 & 0.859 & 0.887 & 0.870 & 0.887 \\ 
PCam/test (10\texttimes) & 0.824 & 0.874 & 0.834 & 0.820 & 0.874 & 0.858 & \textbf{0.876}\\ 
\bottomrule
\end{tabular}}
\label{tab:mpps}
\end{table*}

\subsection{Starting from FMs pre-trained on ImageNet yields faster convergence}
\label{experiment-scratch-vs-pretrained}
\begin{figure}[h]
    \centering
    \includegraphics[width=.48\textwidth]{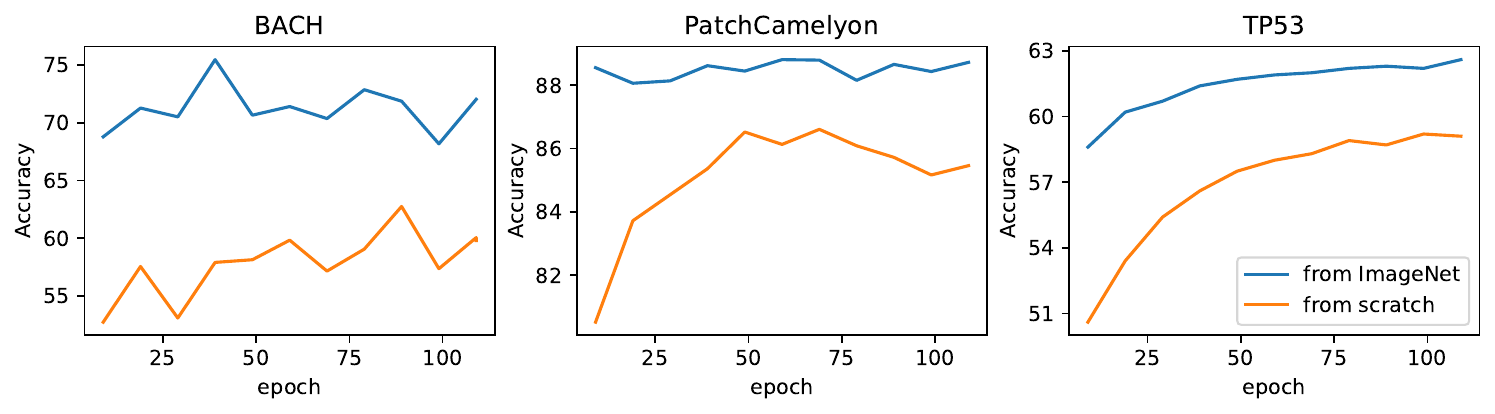}
    \caption{Validation performance over the course of training a ViT-S16 initialized with random weights (blue) and from a model pre-trained on ImageNet (orange) with DINO. Left: Linear probing performance on BACH. Center: Linear probing performance on PCam, Right: Linear probing performance on TP53.}
    \label{fig:scratch-vs-imagenet}
\end{figure}

We seek to leverage the advantages of publicly available pre-trained models in conjunction with domain-aligned pre-training. To investigate this, we assess the impact of initializing from models pre-trained on ImageNet compared to starting from scratch. We use the DINO ViT-S16 architecture with default parameters and train it for 120 epochs. The results are shown in Fig.~\ref{fig:scratch-vs-imagenet}.

Our findings indicate that initializing the FM from pre-trained weights accelerates its convergence and improves the computational efficiency. While training from scratch shows a gradual improvement in performance over the training, it does not converge as fast as fine-tuning a pre-trained model. We hypothesize that initializing from pre-trained weights allows the model to prioritize intricate details within image patches, which is particularly crucial for medical images. As a result, it may be able to converge to a higher level than the models initialized from random weights. A similar effect has also been observed in other works, e.g., in~\cite{DBLP:journals/corr/abs-1912-11370}. We expect to re-evaluate this effect on larger datasets in the future.

\subsection{Training FMs at multiple magnifications improves robustness}
\label{experiment-mpps}

Analyzing pathology images often requires adjusting magnification levels to suit specific contextual demands across different tasks. Lower magnification aids in capturing the overall tissue context, which is particularly beneficial for tasks such as grading prostate cancer. Conversely, tasks focused on individual cell classification benefit from higher magnification to achieve finer resolution. Thus, an ideal pathology FM should be applicable on a range of magnification levels for diverse tasks. We, therefore, introduce patches of various magnification levels during training in the hope that this will enhance the model's versatility and performance across a wider range of downstream tasks, as in~\cite{Kang_2023_CVPR}. We evaluate the impact of training an FM using various magnifications, both individually and mixed. For this purpose, we employ a randomly initialized ViT-S16 and train it on TCGA with DINO for 100 epochs. In addition to the resolutions commonly used in the literature, namely, 40$\times$ and 20$\times$, we include two additional resolutions, 10$\times$ and 5$\times$. The results are presented in Table~\ref{tab:mpps}.

Our analysis reveals that the model trained exclusively on the 20$\times$ magnification level surpasses all other models trained on individual magnifications. However, it does not perform as well as models that simultaneously incorporate multiple magnifications. Our benchmark datasets include various magnification levels, highlighting the lack of consistency in performance across different magnifications for single models, except for the one integrating all four. This integrated model's capability to understand patterns across multiple magnifications provides it with a significant advantage, leading to superior results compared to models trained and evaluated solely on a single magnification level. These findings affirm that we can develop a magnification-agnostic FM without the need for more complex model architectures.

It is also worth noting that mixing patches at multiple magnifications effectively increases the data size and its diversity. The improvement of the model performance, as a result, agrees with our general observations of improved model performance with increasing the data size as discussed in Section~\ref{experiment-data-sizes}.

\label{experiment-data-sampling}
\subsection{The effect of training data size}
In this experiment, we investigate how the performance of the FM changes with scaling up the size of the training data in two different ways: 1) increasing the number of training WSIs and 2) increasing the number of patches sampled from a fixed set of WSIs. Note that in these experiments, we only worked with the TCGA FFPE slides ($\sim$10k WSIs).

\begin{figure}[H]
    \centering
    \includegraphics[width=0.48\textwidth]{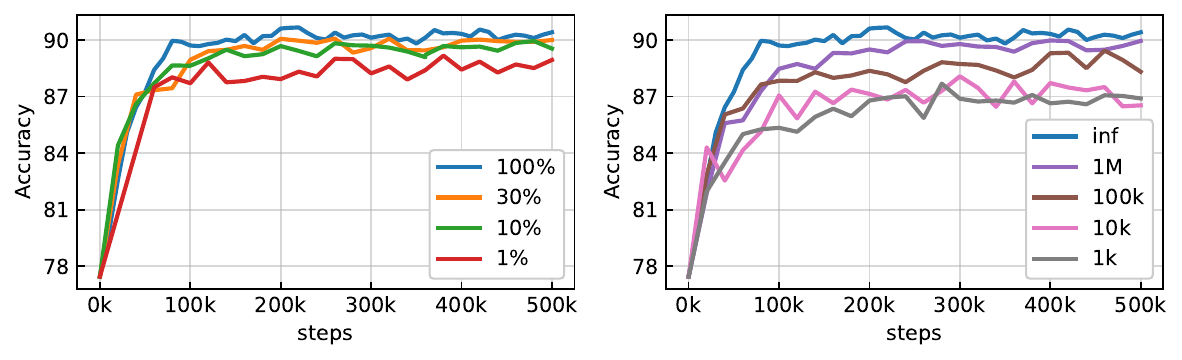}
    \caption{Validation performance of ViT-S16 throughout the DINO training for 100 epochs on the full TCGA dataset and its random 1\%, 10\%, 30\%, and 100\% subsets of WSIs (left) and for different numbers of distinct training patches sampled at random coordinates from random WSIs of 100\% TCGA (right). 'inf' represents the training where all training patches are distinct and are sampled from random coordinates. The performance is measured with linear probing on the PCam/val downstream task.}
    \label{fig:training_data_size}
\end{figure}

\subsubsection{The number of training WSIs}
\label{experiment-data-sizes}

To evaluate how the number of WSIs in the training dataset affects the final performance of the FM, we trained the ViT-S16 model with DINO on random subsets of TCGA of increasing size (1\%, 10\%, 30\%, and 100\% of WSIs randomly sampled from TCGA).
The training was done from scratch on a single GPU for 100 epochs (5000 steps/batches per epoch, with a batch size of 256 patches) and otherwise standard DINO training parameters.

The improvement of the validation accuracy (on the PCam/val downstream task) throughout the training is shown in Fig.~\ref{fig:training_data_size} (left), and the final test accuracy of the trained FMs on the selected downstream tasks is shown in Table~\ref{tab:training_data_size}.
\begin{table}[h]
\centering
\caption{The performance of the ViT-S16 model trained with DINO for 100 epochs on 1\%, 10\%, 30\%, and 100\% subsets of TCGA, as well as the performance of the model initialized with random weights.
The performance is measured as the balanced accuracy of Linear probing on the downstream tasks.
\label{tab:training_data_size}}
\begin{tabular}{lccccc}
\textbf{Training set} & \textbf{BACH} & \textbf{CRC} & \textbf{MHIST} & \textbf{PCam} & \textbf{TP53} \\[0.1ex]
\hline & \\[-1.7ex]
no training           & 0.410         & 0.689        & 0.500          & 0.728         & 0.500         \\
TCGA 1\%              & 0.668         & 0.908        & 0.731          & 0.871         & 0.560         \\
TCGA 10\%             & 0.662         & 0.928        & 0.712          & 0.887         & 0.592         \\
TCGA 30\%             & 0.733         & 0.924        & 0.752          & 0.897         & 0.610         \\
TCGA 100\%            & 0.723         & 0.927        & 0.752          & 0.899         & 0.621\\ 
\bottomrule        
\end{tabular}
\end{table}
The results show that training an FM on more data generally leads to its better performance.
However, the gain diminishes with more training data.
Surprisingly, even with as few as 108 training WSIs (1\% of TCGA), the trained FM provides reasonably high accuracy on all downstream tasks, and the model trained on the 30\% subset is nearly indistinguishable from the FM trained on the full TCGA on all downstream tasks except TP53.

Unlike BACH, CRC, MHIST, and PCam, the TP53 task is TCGA-based and contains WSIs from the same distribution as those used when training the FM.
Note that the DINO SSL training does not see the TP53 labels; hence, this can still be considered a performance on test. We also tested the accuracy on a hold-out subset of TCGA that was not used by DINO and obtained similar results and conclusions, which we skip here for simplicity. 

Based on these results, we believe that 1) FM benefits from more unique samples primarily on in-distribution (ID) data, as demonstrated by the improved performance on on the TP53 task with more training slides; 2) It is necessary to collect more diverse datasets (data from different hospitals, more tissue types, more cancer types, etc.) for the FM to generalize better on out-of-distribution (OOD) data.

\subsubsection{The number of distinct patches}
In this experiment, we limit the number of distinct patches cropped from the WSIs during training.
More precisely, we execute the usual training pipeline with Online Patching (see Section~\ref{sec:online_patching}) and cache the sampled patches on the local hard drive.
After a certain number of patches has been sampled, we randomly sample all the next patches in training from those cached on the local hard drive.

\begin{table}[t]
\centering
\caption{The performance of the ViT-S16 model trained with DINO for 100 epochs on different numbers of distinct training patches.
The patches are cropped from random coordinates at random WSIs of 100\% TCGA.
The first row corresponds to the initial untrained model with random weights.
The last row ('inf' patches) represents the training where all training patches are distinct and are sampled from random coordinates. The performance is measured as balanced accuracy of Linear probing on the downstream tasks.
\label{tab:num_training_patches}}
\begin{tabular}{lccccc}
\textbf{\#\;patches} & \textbf{BACH} & \textbf{CRC} & \textbf{MHIST} & \textbf{PCam} & \textbf{TP53} \\[0.1ex]
\hline & \\[-1.7ex]
$0$          & 0.410         & 0.689        & 0.500          & 0.728         & 0.500         \\
$10^3$       & 0.661         & 0.927        & 0.775          & 0.846         & 0.529         \\
$10^4$       & 0.699         & 0.932        & 0.780          & 0.860         & 0.542         \\
$10^5$       & 0.644         & 0.938        & 0.790          & 0.864         & 0.573         \\
$10^6$       & 0.683         & 0.926        & 0.743          & 0.898         & 0.611         \\
inf          & 0.723         & 0.927        & 0.752          & 0.899         & 0.621         \\ 
\bottomrule        
\end{tabular}
\end{table}

The validation performance (on PCam/val) throughout the training is shown in Fig.~\ref{fig:training_data_size} (right), and the final test accuracy of the trained FMs on the selected downstream tasks is shown in Table~\ref{tab:num_training_patches}.
The first row in Table~\ref{tab:num_training_patches} corresponds to the initial ViT-S16 model with random weights (before training). The last row ('inf' patches) represents the training without restricting the number of distinct patches. Namely, every training patch is sampled from a random WSI at its random position without caching it on the local hard drive, which results in approximately $100\cdot 5000\cdot 256\approx 10^8$ distinct training patches.

Similar to the previous experiment with training the model on subsets of WSIs, we see that even training the model on as few as 1,000 random patches (which is about one patch per ten WSIs) already achieves a reasonable performance on the OOD downstream tasks, and further increasing this number does not lead to drastic improvements of the performance on the OOD tasks.
However, for the in-distribution (ID) TP53 task, the performance grows steadily with exponentially increasing the number of training patches, which suggests that training the FM on more distinct training patches generally leads to better performance, especially on ID data.

The results of these two experiments provide strong evidence that 1) FM training benefits from more unique samples at both slide and patch level; 2) the performance of the FM on OOD tasks can only be significantly improved by substantially enriching and diversifying the training dataset.
Even such a seemingly diverse dataset as TCGA is quickly exhausted in its capacity to facilitate the FM in its ability to generalize on OOD data, highlighting the necessity to go beyond TCGA.

\section{Conclusion}
In this work, we introduced our scalable pipeline for training and evaluating FMs on large pathology imaging data.
Towards building a large-scale pathology model, we developed an online patching technique designed to eliminate the space overhead required to store the patches generated offline.
Through our experiments on TCGA, we demonstrated that online patching does not compromise model performance and may even offer an advantage by providing more diverse data. Furthermore, this technique enables efficient and flexible experimental setup, which could lead to the discovery of novel strategies for training better pathology FMs. These encouraging results allow us to easily scale up our FM training to datasets orders of magnitude larger than TCGA. 

Our experiments on TCGA suggest the following. First, fine-tuning an FM pre-trained on ImageNet on pathology data is more efficient than training a pathology FM from scratch. The initial knowledge contained in the pre-trained FM appears to be relevant for the pathology FM. Second, pathology FMs trained on data of mixed magnifications show a better performance than FMs trained on data of a single magnification. This was, to some extent, anticipated but not fully verified.
This also suggests that, more generally, an FM trained on data of mixed distributions (e.g., data of different magnification or data with different staining), could perform as well as an FM trained on individual distributions separately and provides evidence that an FM could truly be foundational and work well on data from multiple distributions.

We observed clear benefits in scaling up the data size in training ViT-S16. However, only limited benefits were observed when the model size was scaled up. We hypothesize that TCGA in its whole is still not large enough. For example, it was shown in DINOv2~\cite[Fig.~4]{oquab2023dinov2} that the benefits of scaling up the model are more prominent on the larger dataset LVD-142M than the smaller ImageNet-22K dataset.

Similarly, we only observed limited benefits of using DINOv2 compared to DINO on TCGA (Appendix \ref{experiment-DINO-vs-DINOv2}).  
This could be because TCGA is too small to benefit from using the more advanced DINOv2 algorithm or that the downstream tasks we use to evaluate the FMs are not challenging enough to reveal the difference. We leave the analysis for data sizes beyond the TCGA scale for future work.

Through these extensive experiments and analysis, we recognize the importance of a reliable and fair evaluation. We introduced an unsupervised metric off-diagonal correlation that does not require labels on the downstream data and could provide complementary information about the models in addition to the supervised metrics. Finally, we presented our evaluation framework \eva to ensure a consistent evaluation protocol when comparing different FMs.
It is our hope and expectation that other practitioners in the field of medical machine learning and computational pathology contribute new clinically relevant downstream tasks to \eva and adopt it for evaluating their own pathology FMs to ensure the results are comparable across different studies.

We are still at the very beginning of developing a truly foundational pathology FM. It will be worth revisiting the analysis in this work when we scale up the model and data sizes.

\section{Methods}

\subsection{Data}
\label{section:data}
In our experiments, we used collections of WSIs from diverse human tissues across various medical conditions. We describe these datasets below and summarize them in Table~\ref{tab:datasets}.
For the exact partitioning of the datasets into the training, test, and, where applicable, validation subsets, refer to Supplementary Data~\ref{sec:supp}.
\begin{table*}[t]
\caption[Downstream dataset description]{
Summary of benchmark datasets used for linear probing evaluation of FMs.
(*) For the TP53 task, we randomly sampled 102,400 patches from TCGA and assigned the respective TP53 labels derived from their originating WSIs.}
\centering
\scalebox{1.2}{
\begin{tabular}{lccccc}
\textbf{Dataset} & \textbf{\#\;patches} & \textbf{Patch size} & \textbf{Magnification (mpp)} & \textbf{Task} & \textbf{Tissue type} \\[0.1ex]
\hline & \\[-1.7ex]
BACH & 400 & 1536\texttimes2048 & 20\texttimes \, (0.42\,µm/px)  & Cls (4 classes) & Breast \\ 
CRC & 107,180 & 224\texttimes224 & 20\texttimes \, (0.50\,µm/px)  & Cls (9 classes) & Colorectal \\ 
MHIST & 3,152 & 224\texttimes224 & 5\texttimes \, (2.00\,µm/px)
& Cls (2 classes) & Colorectal Polyp \\
PCam & 327,680 & 96\texttimes96 & 10\texttimes \, (0.97\,µm/px)
& Cls (2 classes) & Breast lymph node\\
TP53 & 102,400* & 224\texttimes224 & 20\texttimes \, (0.50\,µm/px)  & Cls (2 classes) & All TCGA tissues \\ 
\bottomrule
\end{tabular}}
\label{tab:datasets}
\end{table*}

\textbf{TCGA} This dataset contains approximately 29k hematoxylin and eosin (H\&E) stained tissue slides from 32 cancer types at different microscopic magnifications, collected at different hospitals for The Cancer Genome Atlas (TCGA) project~\cite{Chang:2013aa} by the TCGA Research Network: \url{https://www.cancer.gov/tcga}.
TCGA has been widely used for training foundation models on pathology images~\cite{Filiot2023.07.21.23292757,Kang_2023_CVPR,chen2022scaling}.
Following these efforts, we used this dataset to train our foundation models.

\textbf{TCGA TP53} 
From TCGA metadata, we constructed a downstream task for predicting TP53 status from WSIs in TCGA.
To this end, we assess TP53 to be aberrated when it either harbors a mutation or when both copies are deleted. 
Following \cite{Donehower2019-tp53}, we consider all mutations that are either labeled as moderate (e.g., non-synonymous missense) or high (e.g., nonsense, frameshift) without filtering on variant allele frequency (VAF). 
The rationale for including non-synonymous missense mutations is that these mutations nearly always showed a high VAF, suggesting positive selection pressure and, hence, functional impact \cite{Donehower2019-tp53}.
The rationale for not filtering on VAF (for the remaining mutations, labeled as high impact) is that for nearly all cases where only one allele is mutated, a second mutation could be determined, suggesting the other allele has also been disabled \cite{Donehower2019-tp53}. 
Following this approach, we identified roughly 6k tumors with functional TP53 and roughly 3.5k tumors with non-functional TP53 in TCGA. We have made these data available for download (see \ref{sec:supp}).

The TP53 status is a patient-level signal; however, in this work, we treat it as a patch-level signal. That is, the task is to predict TP53 status from a patch of a WSI instead of the whole slide. This introduces label noise, as the WSI-level signal will most likely not be detectable from every patch within the WSI. Also, there could be heterogeneity in the expression of the involved genes. Nonetheless, we find that this task can be used to compare between FMs and could be a valid metric. We leave the construction of the slide-level metric for future work.

For evaluation, we randomly sample 102,400 patches from all the TCGA diagnostic slides with equal probability for each slide to be sampled, and we report linear probing balanced accuracy on 5-fold cross-validation of patient-based splits of the patches.

\textbf{BACH} This dataset contains 400 images originally generated for the Grand Challenge on BreAst Cancer Histology images challenge~\cite{bach_dataset}.
Each image is of size 1536$\times$2048 pixels, at a scale of 0.42\,µm/pixel, and belongs to one of four classes: 1) normal,
2) benign, 3) {\it in situ} carcinoma, and 4) invasive carcinoma, with each of the four classes having exactly 100 images assigned to it.
We downloaded all images with their corresponding metadata from \url{https://iciar2018-challenge.grand-challenge.org} and used this metadata to split the entire dataset into a training and a test part, such that different images from the same patient never appear in both training and test parts but only in one of them.
As a result, the test set contains 132 images, with 23, 48, 30, and 31 images from each of the four classes, respectively, which is roughly one-third of the entire dataset. Note that this differs from the existing literature, where typically a random split is performed without grouping the patches by patient, e.g., in~\cite{Kang_2023_CVPR}.
For the exact composition of the training and test sets, refer to Supplementary Data~\ref{sec:supp}.

\textbf{CRC} The CRC dataset \cite{kather_2018_1214456} comprises 100,000 training and 7,180 test images (224\texttimes224 pixels) at 20\texttimes \, magnification, sourced from H\&E stained WSIs representing human colorectal cancer and normal tissue. The training set is derived from 86 WSIs, while the test set is sourced from 25 WSIs. These WSIs are obtained from the NCT Tissue Bank and the University Medical Center Mannheim. The objective is to classify nine tissue classes: adipose tissue, background, debris, lymphocytes, mucus, smooth muscle, normal colon mucosa, cancer-associated stroma, and CRC epithelium. All images undergo color normalization using the Macenko method (NCT-CRC-HE-100K). We do not make any use of the unnormalized (NCT-CRC-HE100K-NONORM) variants.

\textbf{PatchCamelyon (PCam)} This dataset consists of 327,680 patches of 96$\times$96 pixels at a FoV of 0.97\,µm/px~\cite{Veeling2018-qh}. These patches are cropped from WSIs of breast lymph node sections and are marked with binary labels indicating the presence of metastatic tissue in the image.

\textbf{MHIST} The MHIST dataset \cite{MHIST} consists of 3,152 H\& E-stained FFPE fixed-size images (224\texttimes 224 pixels) of colorectal polyps, where each image is assigned to one of two classes: 1) Hyperplastic Polyp (HP) or 2) Sessile Serrated Adenoma (SSA).

\textbf{CoNSeP}
The Colorectal Nuclear Segmentation and Phenotypes (CoNSeP) dataset~\cite{graham2019hover} consists of 41 H\&E 1000$\times$1000 pixel images, and it is split into 27 and 14 images for training and test sets.
The data comes from the University Hospitals Coventry and Warwickshire, UK. 
The annotation contains segmentation masks of each nucleus along with the  grouped classes as described in~\cite{graham2019hover}:
1) other, 2) inflammatory, 3) healthy \& dysplastic/malignant epithelial, 4) spindle-shaped. To evaluate FMs, we used the CoNSeP dataset as a semantic segmentation task, where each pixel is assigned to one of the five categories: the four cell-type categories and background. We report the DICE score without background.

\begin{figure}[t]
    \centering
    \includegraphics[width=0.48\textwidth]{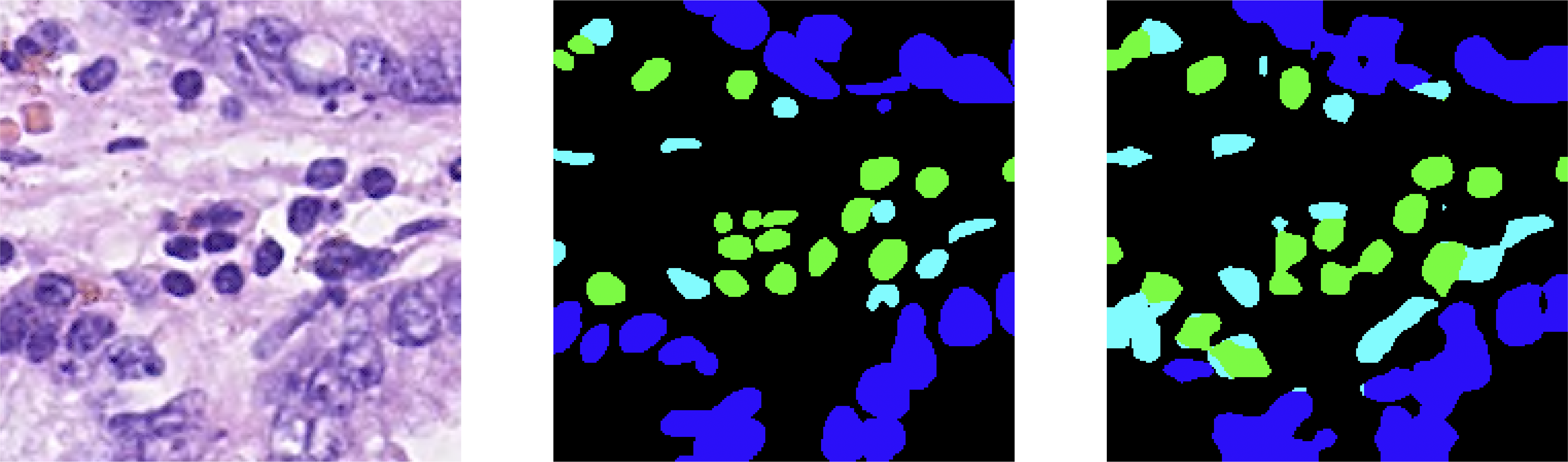}
    \caption{Quantitative results for the semantic segmentation task. Left: the patch. Center: the semantic segmentation labels. Right: the predictions with our ViT-B8 model.}
    \label{fig:segmentation}
\end{figure}

\subsection{Online high-throughput loading of patches from WSIs}
\label{sec:online_patching}
Self-supervised pre-training on WSIs is, as of date, usually not performed at the slide level due to the gigapixel-level size of the WSIs, which exceeds the GPU memory of standard modern hardware. Although sophisticated techniques such as activation checkpointing could be used \cite{pinckaers2020streaming,dooper2023gigapixel}, they usually have a significant performance impact for large models. Therefore, state-of-the-art pathology FMs are currently trained at the patch level, where smaller patches must be extracted. Typically, patches are extracted and stored offline before training to enable the efficient loading of patches during training; see e.g.~\cite{campanella2023computational}. In addition to the significant space overhead required to store the pre-processed image patches, this limits the flexibility in the choice of patch size and the magnification level, as any change in these parameters requires another preparation of the patches.

To allow for more flexibility, we developed {\it Online Patching}, a library that enables high-throughput extraction and loading of patches from WSIs during training. The library allows extracting from any WSI patches at completely arbitrary coordinates and at arbitrary magnification levels. This allows training models on virtually all the patches contained in the WSIs without having to store the patches on disk.

To sample only foreground patches, a U-Net-based foreground segmentation model is used to compute the foreground masks of the WSIs at a lower resolution, usually at thumbnail scale. From this mask, a polygon is computed that can be efficiently stored in memory. During training, a slide is first sampled from all available slides, where different slide-level sampling strategies could be specified (e.g., uniform random sampling or prior-based sampling). From the sampled slide, patch coordinates are randomly sampled with a minimum area overlap of a candidate patch with the polygon. The patch is then extracted from the image level closest to the target magnification level and resized to the target patch size. This method of patch selection increases the diversity of the patches used in training and allows training on more patches than what would be possible with a fixed set of patches. At inference time, patches with sufficiently many foreground pixels can be generated by iterating a grid of a specified size.

The online patching library utilizes a virtual in-memory filesystem to make the whole of TCGA accessible as a single \textit{Zarr} data source \cite{alistair_miles_2024_10790679}. The virtual filesystem allows the original SVS files to be accessible as if they were stored in the Zarr data format, requiring only minimal extraction of the tile byte ranges before running online patching. The library provides optimized functionality for the asynchronous loading of tiles from network or blob storage. Related open-source initiatives are being developed to use Zarr as a unified format for biomedical images \cite{moore2021ome,moore2023ome}.

We are aware of other possible solutions where patch extraction is performed on intermediate servers, providing an API to abstract away that complexity. An example of that approach is the WSI DICOMWeb python library from Google \cite{wsi_dicomweb_blog}, which provides a way to extract patches from images stored in the Google DICOM store. The downside of solutions that rely on intermediate servers performing the patch extraction is that it increases the complexity of the infrastructure required (for larger datasets, the patch extraction servers would need to scale up accordingly). Our online patching library, on the other hand, works with WSIs directly and in a cloud-agnostic way.

\subsection{Pretraining setup}
\label{subsection:pretraining setup}
We adhere to the suggested training methodology for natural images as outlined in DINO \cite{caron2021emerging} and DINOv2 \cite{oquab2023dinov2} with slight adjustments: (i) model initialization is performed using ImageNet SSL weights, (ii) the learning rate is reduced by a factor of 10, (iii) random sampling of 256$\times$256 patches is carried out with a minimum of 40\% foreground presence, (iv) training encompasses multiple magnification levels, specifically 5$\times$, 10$\times$, 20$\times$, and 40$\times$, and (v) image normalization is conducted using a non-informative mean and standard deviation of 0.5 to scale values within the [-1, 1] range.

In all our experiments, we adopt the ImageNet Epoch concept \cite{kang2023benchmarking} and define one epoch as 1,280,000 patches.

The pre-training occurred in two phases: In the first phase, we trained exclusively on FFPE slides for 100 ImageNet epochs. In the second phase, we extended the training by another 100 epochs, incorporating the FF slides while reducing the peak learning rate by half compared to the initial stage.

The ViT-B8 was trained on 8 H100 GPUs with a batch size of 32 per GPU. The DINOv2 ViT-L14 was trained on 16 H100 GPUs with batch size per GPU 32, and all other our models from Table~\ref{tab:results} were trained on 4 H100 GPUs and batch size per GPU 256 for ViT-S16, 64 for ViT-S8 and 128 for ViT-B16.

\subsection{Evaluation setup}
To evaluate different training strategies, we apply the trained FMs on a selection of downstream tasks using public datasets to generate embeddings given input images. 
The performance of the FMs is evaluated with two groups of metrics: 1) metrics that evaluate the quality of the representations directly without labels; 2) metrics measuring the performance of the representations on the downstream prediction tasks with a lightweight head network, where labels are necessary.

\subsubsection{Unsupervised metrics}
\label{sec:unsupervised metric}
To evaluate FMs without the need for labeled data, we use RankMe \cite{garrido2023rankme} as one of the metrics in this study. RankMe estimates the rank of embeddings of test data, and it has been shown to correlate well with downstream performance.

In addition, we introduce another simple unsupervised criterion to evaluate the quality of the representations directly: off-diagonal correlation (\odcorr), which simply measures the average correlation coefficient between the embeddings of different samples in the evaluation dataset, i.e., the off-diagonal elements of a correlation matrix. This metric is motivated by the observation that when the embeddings of different samples are different enough and, thus, not correlated, the samples can be distinguished from each other based on their embedding vectors. This is necessary for learning downstream tasks.  

\odcorr is calculated based on the 2D matrix of embedding dimensions and samples, and it takes the square root of the mean square of the correlations between embedding vectors of pairs of \textit{samples}. In related earlier work, the correlations over this 2D matrix are computed in the orthogonal direction, i.e., the correlations between sample-value vectors of pairs of \textit{embedding dimensions} are considered. 
\cite{cogswell2015reducing} showed that the amount of correlation in hidden activations corresponds with the amount of overfitting, and \cite{zhou2021isobn} visualized the Pearson correlation coefficient of [CLS] embeddings such that highly correlated dimensions are located near each other in blocks. 

The off-diagonal correlation metric ranges between $0$ and $1$, with $0$ indicating no correlations between samples and $1$ indicating all samples are correlated with each other. Since this metric simply measures the correlations between different samples, it can be used to compare models of different dimensions. Denoting the embedding matrix of an evaluation dataset as $\bm{Z}$ of shape $N \times K$, the off-diagonal correlation is formally defined as:
\begin{equation}
    \odcorr(\bm{ Z} ) = \sqrt{\frac{\sum_{i\neq j}\rho(Z_i, Z_j)^2}{N(N - 1)}},
\end{equation}
where $\rho(Z_i, Z_j)$ is the Pearson correlation coefficient for the embeddings of samples $i$ and $j$.

In Appendix, Section~\ref{supp:odcorr}, we show that \odcorr highly correlates with the downstream performance and can be compared between models of different sizes.

\subsubsection{Evaluation framework: \textit{eva}}
\label{section:eva}

To evaluate foundation models on out-of-distribution (OOD) downstream tasks, we use our open-source evaluation framework \textit{eva}, which has been designed to provide an explainable, fair, and easily reproducible FM-evaluation standard across backbone sizes and architectures.

\textit{eva} aims to support a large selection of public datasets and tasks. In the first release, PCam~\cite{Veeling2018-qh}, BACH \cite{bach_dataset} for breast cancer classification and colorectal (CRC) cancer classification \cite{kather_2018_1214456} are included. To evaluate an FM on a downstream task, \textit{eva} prepares a task dataset, performs inference to compute the embeddings, trains a head model for that task, and evaluates the performance.

If a dataset has a designated validation or test split, we use it for evaluation and report results accordingly. However, if such splits are not available, we create a stratified split to ensure proper separation of slides or patients, thus preventing any potential data leakage.

\textit{eva} prioritizes meaningful evaluation of the FMs over maximum individual downstream performance. Therefore, the head architecture is deliberately chosen to be lightweight, robust in downstream performance, and with minimal bias toward any particular FM architecture.

To achieve this, \textit{eva} follows a standard protocol introduced in~\cite{vorontsov2024virchow} that trains a single-layer MLP with a fixed number of training steps and hyperparameters. For small datasets, we reduced the batch size and linearly scaled down the learning rate. To prevent overfitting, \eva applies early stopping after 5\% of the maximum number of epochs.
For a detailed configuration, see Appendix, Table~\ref{tab:head_setup}.
We found that with this setup, 
we achieve stable results across multiple runs for each of the evaluated tasks and FMs.

For the semantic segmentation evaluation on the CoNSeP dataset, we trained on randomly cropped patches of size 224 from the training set and evaluated on grid patches of the same size with stride 194 on the test set. We used Mask2Former \cite{Mask2Former} as a decoder on top of the FMs. We kept the FM frozen and the decoder light. We deliberately reduced the capacity of the decoder (number of queries: 32, number of encoder layers and attention heads: 4, feature size and hidden dimension: 32). In the same spirit of keeping a lightweight decoder, we did not use ViT-Adapter~\cite{ViT-Adapter}, which is sometimes used when evaluating FMs, e.g.,~in \cite{chen2023generalpurpose}. We observed some instabilities in training the segmentation decoder. Not all training runs converged in the maximal number of epochs. Thus, for each FM, the results are averaged over three runs that did converge.

\section{Supplementary Data}
\label{sec:supp}
The model checkpoints, and the information for reproducing the evaluation results presented in this work are available for download from \repourl.

\section{Acknowledgments}
The authors thank Jonas Teuwen, Eric Marcus and the AI for Oncology group at the Netherlands Cancer Institute (NKI) for the fruitful discussions and collaborations. They have also kindly provided a segmentation annotation dataset, which was used to train the foreground/background segmentation model that identifies the foreground regions used by the Online Patching method described in this article.

\bibliographystyle{ieeetr}
\bibliography{bibliography}

\newpage

\appendix

\subsection{Evaluation setup}
\label{appendix_eval}

For every experiment conducted, we adhered to the setup described in the evaluation of Virchow \cite{vorontsov2024virchow}. Specifically, we employed a linear projection classifier with a batch size of 4,096, utilizing the stochastic gradient descent (SGD) optimizer with a cosine learning rate schedule ranging from 0.01 to 0 over 12,500 iterations. This was done on top of the embeddings produced by the frozen encoder. Moreover, we implemented early stopping, halting training after 5\% of the total training epochs. For further details, refer to Table \ref{tab:head_setup}.

\begin{table}[bp]
\caption[Head setup]{Hyperparameters for the head used in downstream evaluation in \eva.}
\centering
\begin{tabular}{lccccccccc}
\toprule
Backbone & frozen \\ 
Hidden layers & None \\ 
Dropout & 0.0 \\
Activation function & None \\ 
Number of steps & 12,500 \\ 
Base batch size & 4,096 \\ 
Batch size & dataset specific \\ 
Base learning rate & 0.01 \\ 
Learning rate & dataset specific \\ 
Early stopping & [Max epochs] / 20 \\ 
Optimizer & SGD \\ 
Momentum & 0.9 \\ 
Weight Decay & 0.0 \\ 
Nesterov momentum & True \\ 
LR Schedule & Cosine without warmup \\
\bottomrule
\end{tabular}
\label{tab:head_setup}
\end{table}

\subsection{No significant benefits of DINOv2 over DINO on TCGA}
\label{experiment-DINO-vs-DINOv2}
Kang et al \cite{Kang_2023_CVPR} have concluded that "there is no clear winner" in different SSL methods they have examined, including MoCo v2 \cite{mocov2-DBLP:journals/corr/abs-2003-04297}, SwAV \cite{swav-DBLP:journals/corr/abs-2006-09882}, Barlow Twins \cite{barlow-twins-DBLP:journals/corr/abs-2103-03230} and DINO \cite{caron2021emerging}. However, the ViT models trained with DINO did seem to be superior in label efficiency (in \cite[Table 6]{Kang_2023_CVPR}). 

Filiot et al \cite{Filiot2023.07.21.23292757} have compared iBOT \cite{zhou2021ibot} with MoCo v2 \cite{mocov2-DBLP:journals/corr/abs-2003-04297} and shown better performance with iBOT. But the comparison with DINO was not conclusive as the DINO model was only trained on one specific cohort instead of on the whole pan-cancer dataset.

DINOv2 \cite{oquab2023dinov2} has been introduced to incorporate the best of DINO and iBOT algorithms. In Table 1 of the DINOv2 paper \cite{oquab2023dinov2} the benefits of different components in the DINOv2 algorithm in comparison to the iBOT algorithm \cite{zhou2021ibot} were shown. However, no such comparison has been made between DINO and DINOv2. In comparison to DINO \cite{caron2021emerging}, DINOv2 introduced the following three main components among other improvements:
\begin{itemize}
    \item Patch-level objective, and untied head weights between image- and patch-level objectives
    \item Sinkhorn-Knopp centering
    \item KoLeo regularizer
\end{itemize}

The patch-level objective, in particular, increases the complexity of loss computation and requirements on GPU memory by the number of patches per image. It is not clear a priori whether this increased complexity is justified by, e.g., greater data efficiency or other benefits, especially on smaller data and model sizes. 

In our experiments, we did not observe significant benefits of models trained by DINOv2 over DINO on TCGA. As shown in Table~\ref{tab:results}, a larger model (ViT-L14) trained with DINOv2 only achieved comparable performance in downstream tasks as the smaller ViT-S16 model trained with DINO. 

\begin{figure*}
    \centering
    \includegraphics[width=\textwidth]{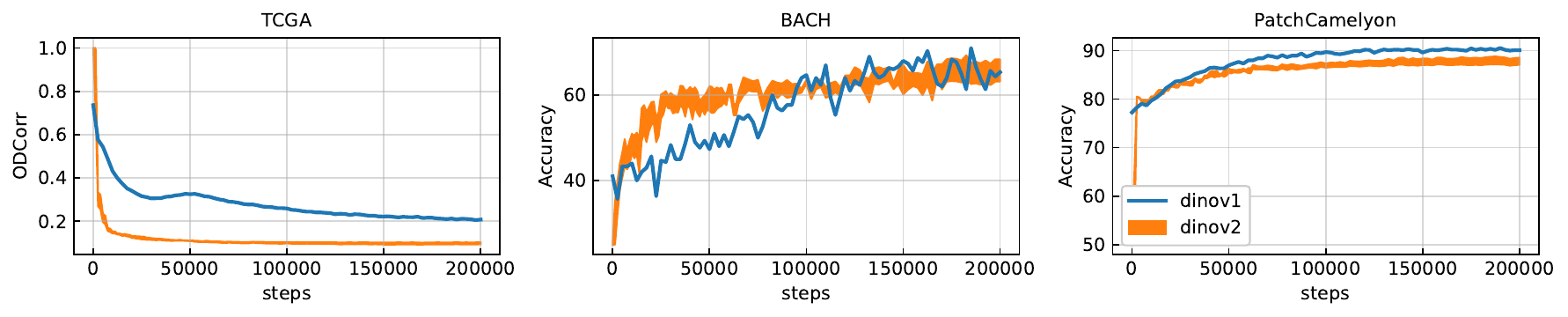}
    \caption{Validation performance over the course of training a ViT-S16 model using DINOv2 (orange) and DINO (blue). Left: Off-diagonal correlation on randomly selected TCGA patches. Center: Linear probing performance on test split of BACH dataset. Right: Linear probing performance on validation split of PCam dataset. The orange curves for DINOv2 show a range from 4 different runs with different learning rates, while the blue curves show one single run with DINO using the standard setting, details can be found in Section~\ref{appendix_eval}.}
    \label{fig:DINOv1_vs_v2}
\end{figure*}

Furthermore, we also examined the individual learning curves with the two algorithms to check whether one learns faster than the other (detailed setup in Section~\ref{sec:supp}). Fig.~\ref{fig:DINOv1_vs_v2} shows the learning curves of training a ViT-S16 model using DINOv2 and DINO algorithms on a single A100-80GB machine, as measured by the linear probing performance on OOD datasets: BACH and PCam, as well as the off-diagonal correlation on in-distribution TCGA data. We have observed that DINOv2 tends to take more time for one training step than DINO. However, we do not compare by walltime directly here as this could be affected by many other factors, such as latency in connecting to cloud storage, amount of validation during training, etc. The number of training steps, on the other hand, is comparable, as we use the same batch size of 256 across different experiments.

No significant difference can be observed in the BACH and PCam learning curves between the two learning algorithms. The models trained with DINOv2 reached comparable performance in comparable number of training steps as the models trained with DINOv1. 

However, we do observe better off-diagonal correlation with DINOv2 trained models. We hypothesize that the DINOv2 trained models do tend to generate more distinguishable embeddings, as demonstrated by the left panel in Fig.~\ref{fig:DINOv1_vs_v2} and in Table \ref{tab:results} and its extended version Table \ref{tab:results-appendix}. The downstream prediction task may however not necessarily need this level of distinction between samples, thus the benefits do not necessarily show up in downstream tasks.

In summary we believe that DINOv2 may be superior to DINOv1 at the expense of slightly more compute and resource requirements. The benefits, however, may not directly translate into downstream task performance, especially if the tasks are relatively simple.

\subsubsection{Pretraining setup in comparing DINO vs DINOv2}
To compare DINO with DINOv2 algorithm, a ViT-S16 model was trained with both training algorithm on one A100-80GB machine with the following hyperparameters in Table~\ref{tab:setup-dino-dinov2}:

\begin{table}[t]
\caption[DINO-DINOv2 pre-training details]{
Summary of hyperparameters for training in the experiment comparing DINO and DINOv2 (Section \ref{experiment-DINO-vs-DINOv2}).}
\centering
\begin{tabular}{lcc}
 & \textbf{DINO} & \textbf{DINOv2}  \\[0.1ex]
\hline & \\[-1.7ex]
Batch size & 256 & 256\\ 
Learning rate & 0.0005 & 0.0005, 0.001, 0.002\\ 
Steps per epoch & 5,000 & 5,000 \\ 
Max epochs & 100 & 100\\ 
iBOT separate head & n/a & False \\
Layer wise decay & n/a & 0.9, 1\\
\bottomrule
\end{tabular}
\label{tab:setup-dino-dinov2}
\end{table}

Note that this is not the same setting as we used for training the models in Table~\ref{tab:results}, as here in order to evaluate the effect of DINOv2 we have kept everything else the same except for the loss definition and the data transformation.

\subsection{\odcorr correlates with downstream performance}
\label{supp:odcorr}
Losses in various SSL algorithms are usually not very informative, and, in particular, they are usually not indicative of the performance of the FMs on the downstream tasks. One way to evaluate the quality of the FMs is to apply it on some labeled datasets and evaluate the performance on the downstream tasks. However, the evaluation is constrained, and/or biased by the labels available. A few metrics have been proposed to address this, such as RankMe \cite{garrido2023rankme} which estimates the embeddings' rank or $\alpha$-ReQ \cite{ghosh2022investigating} which estimates the eigenspectrum decay of the representations.

In \cite{garrido2023rankme} it is claimed that RankMe can consistently predict downstream performance for linear and non-linear probing, however, as RankMe depends on the dimension of the representations, it is not comparable between models of different embedding dimensions and should "only be used to compare different runs of a given method". In this study we use RankMe as one of the metrics to evaluate the different training strategies.

\subsubsection{On natural images}
To evaluate the effectiveness of the \odcorr metric in general, we evaluate public available trained models from timm \cite{rw2019timm} on a few public datasets (CIFAR10 and CIFAR100 \cite{cifar} and Food101 \cite{bossard14}. In particular we have chosen all pre-trained vit-small models that are available in the timm library which are 17 pre-trained models using a wide range of algorithms from supervised to self-supervised pre-training. Together with the 3 public datasets on natural images, they provide a diverse testing ground to evaluate quality of the \odcorr metric.

Following the same protocol as in \cite{garrido2023rankme} we train a linear head on the frozen backbone on the train split of the dataset and compute the top-1 accuracy and the \odcorr on the test split. 

\begin{figure*}[h]
    \centering
    \includegraphics[width=\textwidth]{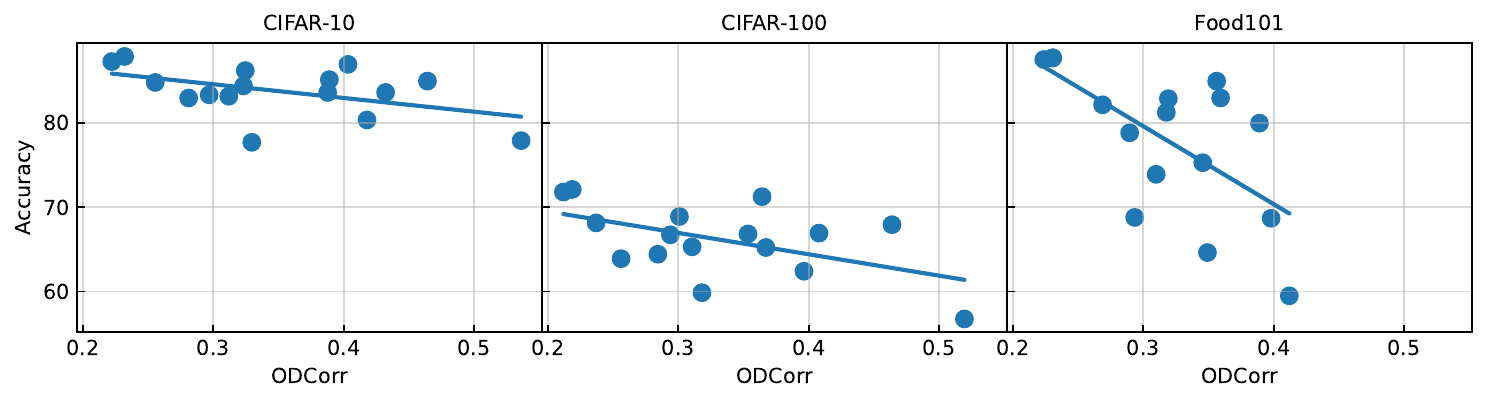}
    \caption{Correlation between \odcorr and top-1 accuracy of the representation on CIFAR-10 (left), CIFAR-100 (center) and Food101 (right). An inverse correlation between the \odcorr and top-1 accuracy can be observed.}
    \label{fig:odcorr-natural-images}
\end{figure*}

As can be seen in Fig.~\ref{fig:odcorr-natural-images} for a given dataset, a lower \odcorr usually corresponds to a higher top-1 accuracy. 

\subsubsection{On whole slide images}
We also evaluate the \odcorr metric on different pathology FMs collected from all the above experiments (e.g., for data sizes, for initialization strategies) regardless of how the FM was trained. In particular, we evaluate the relation between linear probing performance and \odcorr, between linear probing and RankMe, as well between \odcorr and RankMe. The linear probe was trained on the respective train split and the results reported on the test split. The \odcorr and RankMe was calculated on the test split only.

\begin{figure*}[h]
    \centering
    \includegraphics[width=\textwidth]{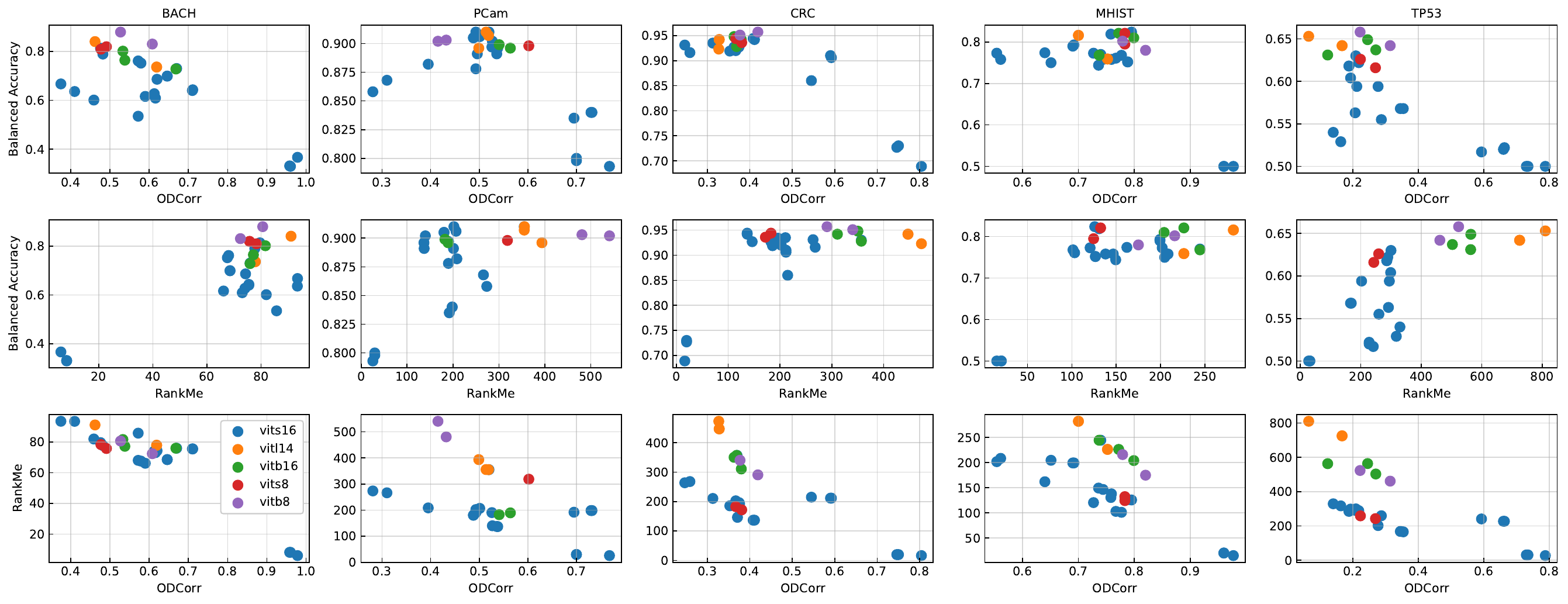}
    \caption{Relation between linear probing performance, \odcorr and RankMe of the representations on pathology datasets: (upper) Balanced Accuracy vs. \odcorr, (middle) Balanced Accuracy vs. RankMe, (bottom) RankMe vs. \odcorr. Models of different sizes are colored differently}
    \label{fig:odcorr-wsis}
\end{figure*}

In Fig.~\ref{fig:odcorr-wsis}, we first observe in the bottom panel that there is an inverse relation between RankMe and \odcorr where lower \odcorr correlates with higher RankMe, as expected. In addition we also observe that there could be different relations between different model sizes as RankMe depends on the model sizes.

Secondly we observe in the top panel also an inverse relation between the linear probing performance and \odcorr across datasets, where lower \odcorr correlates with higher linear probing performance. Interestingly the linear probing performance plateaus when the \odcorr goes below certain value, e.g., for CRC the linear probing performance does not improve anymore once the \odcorr drops below 0.5, similarly for MHIST although the threshold is higher at around 0.8. For BACH and TP53 the trend is not stopped in all our experiments. For PCam there seems to be a peak at \odcorr at 0.5, and the performance drops with further decreasing \odcorr. However, this could be due to that the fact that the \odcorr is calculated on the test split only, and the linear probing performance is also influenced by the quality of the embeddings of the train split. The existence of the plateau highlights a situation where the \odcorr can provide complementary information to the linear probing situation: for the points on the plateaued part of the curves, the linear probing is no longer able to differentiate between the models as they all show similar performance, in the mean time the \odcorr can still be used to identify better models

In the middle panel we observe the same plateau with linear probing vs. RankMe, i.e., after the RankMe reaches certain number, the linear probing performance stops growing further, in agreement with what we observe with \odcorr. On the other hand, as the RankMe is usually higher with higher embedding dimension, we almost always observe the ViT-L14 models on the right end of the curve, suggesting that they are superior. But this is not necessarily the case, as shown in the upper panel, the ViT-L14 models are not always the best at distinguishing samples as demonstrated by the sometimes higher \odcorr values.

In summary, we believe that \odcorr could be a useful unsupervised metric to provide additional information about the FMs, especially when the supervised metrics of downstream tasks do not differ that much anymore. In addition, it could also be useful in comparing models of different sizes.

\begin{table*}[t]
\caption[Patch-level benchmarks]{Linear probing evaluation of FMs on patch-level downstream datasets. We report averaged balanced accuracy over 5 linear probing runs and the DICE score on the foreground pixels for the CoNSeP task. Values from Virchow are taken from \cite{vorontsov2024virchow}. All other results were generated using  \textit{eva} (except CoNSeP that will be soon supported). We compare the performance from a randomly initialized ViT-S16 model (first line), the generic FMs pre-trained on ImageNet (above the dashed line), and the pathology specific FMs (below the dashed line)}
\centering
\scalebox{1.1}{
\begin{tabular}{lccccccccc}
\textbf{Model} & \textbf{BACH} & \textbf{CRC} & \textbf{MHIST}  & \textbf{PCam/val} & \textbf{PCam/test} & \textbf{CoNSeP}\\[0.1ex]
\hline & \\[-1.7ex]
ViT-S16 \color[HTML]{9B9B9B} \textit{(rand. weights)} &0.410 (±0.009) & 0.617 (±0.008) & 0.501 (±0.004) & 0.753 (±0.002) & 0.728 (±0.003) & 0.583 (±0.012)\\ 
DINO ViT-S16 \cite{caron2021emerging} & 0.695 (±0.004) & 0.935 (±0.003) & \uuline{0.831 (±0.002)} & 0.864 (±0.007) & 0.849 (±0.007) & 0.611 (±0.018)\\ 
DINO ViT-B8 \cite{caron2021emerging} & 0.710 (±0.007) & 0.939 (±0.001) & 0.814 (±0.003) & 0.870 (±0.003) & 0.856 (±0.004) & 0.710 (±0.005)\\[1ex]
\arrayrulecolor{gray} \hdashline[1.5pt/1.5pt] \\[-1.5ex]
Lunit \cite{Kang_2023_CVPR} & 0.801 (±0.005) & 0.934 (±0.001) & 0.768 (±0.004) & 0.889 (±0.002) & 0.895 (±0.006) & 0.654 (±0.003)\\ 
Phikon \cite{Filiot2023.07.21.23292757} & 0.725 (±0.004) & 0.935 (±0.001) & 0.777 (±0.005) & \uline{0.912 (±0.002)} & \uline{0.915 (±0.003)} & 0.666 (±0.004)\\
DINO ViT-S16 \color[HTML]{9B9B9B} \textit{(ours)} & 0.797 (±0.003) & 0.943 (±0.001) & \uline{0.828 (±0.003)} & 0.903 (±0.001) & 0.893 (±0.005) & 0.649 (±0.013)\\ 
DINO ViT-S8 \color[HTML]{9B9B9B} \textit{(ours)} & 0.834 (±0.012) & 0.946 (±0.002) & \uuline{0.832 (±0.006)} & 0.897 (±0.001) & 0.887 (±0.002) & \uline{0.724 (±0.007)}\\ 
DINO ViT-B16 \color[HTML]{9B9B9B} \textit{(ours)} & 0.810 (±0.008) & \uuline{0.960 (±0.001)} & 0.826 (±0.003) & 0.900 (±0.002) & 0.898 (±0.003) & 0.658 (±0.011)\\ 
DINO ViT-B8 \color[HTML]{9B9B9B} \textit{(ours)} & \uline{0.865 (±0.019)} & \uline{0.956 (±0.001)} & 0.809 (±0.021) & \uuline{0.913 (±0.001)} & \uuline{0.921 (±0.002)} & \uuline{0.741 (±0.002)}\\ 
DINOv2 ViT-L14 \color[HTML]{9B9B9B} \textit{(ours)} & \uuline{0.870 (±0.005)} & 0.930 (±0.001) & 0.809 (±0.001) & 0.908 (±0.001) & 0.898 (±0.002) & 0.679 (±0.007)\\[1ex]
\arrayrulecolor{gray} \hline \\[-1.5ex]
Virchow \cite{vorontsov2024virchow} & - & 0.962 & 0.830 & - &0.933\\
\bottomrule
\end{tabular}}
\label{tab:results-appendix}
\end{table*}

\end{document}